\documentclass{llncs}

\usepackage{newalg}
\usepackage{times}
\usepackage{graphicx}

\begin{document}



\title{Manipulability of Single Transferable Vote}


\author{Toby Walsh\inst{1}}
\institute{NICTA and UNSW, Sydney, Australia}

\maketitle

\begin{abstract}
For many voting rules, 
it is NP-hard to compute a successful manipulation. 
However, NP-hardness only
bounds the worst-case complexity. Recent theoretical
results suggest that manipulation may often
be easy in practice. We study empirically the 
cost of manipulating the
single transferable vote (STV) rule.
This was one of the first rules
shown to be NP-hard to manipulate. It also appears to be
one of the harder rules to manipulate since it involves
multiple rounds and since, unlike many other rules, it is 
NP-hard for a single agent to manipulate without weights on
the votes or uncertainty about how the other
agents have voted. In almost every election in our 
experiments, it was easy to compute how a single
agent could manipulate the election or to prove
that manipulation by a single agent was impossible. 
It remains an interesting open question if 
manipulation by a coalition of agents is hard
to compute in practice. 
\end{abstract}





\section{Introduction}

A simple mechanism for collaborating agents to
reach consensus on a plan of action is
to vote. One issue is the possibility of
agents trying to manipulate
such an election by mis-reporting their preferences in
order to get a better result. 
Fortunately, it is often
computationally difficult to find a successful
manipulation 
\cite{bartholditoveytrick}.
For example, Bartholdi and Orlin proved
that the Single Transferable Vote
(STV) rule is NP-hard to manipulate \cite{stvhard}.
This remains one of the few commonly used voting rules
which is NP-hard to manipulate {\em without} 
weights on the votes, or uncertainty
in how the other agents have voted. 
NP-hardness is only a worst-case result and may not reflect
the difficulty of manipulation in practice. Indeed, a number of 
recent theoretical results 
suggest that manipulation can often be computationally
easy \cite{csaaai2006,prjair07,xcec08,fknfocs09,xcec08b}. 
Most recently, Walsh has suggested that 
empirical studies might provide insights
into the computational complexity of manipulation
that can complement such theoretical
results \cite{wijcai09}. 
For example, when theoretical analysis is asymptotic,
empirical studies can reveal if hidden constants
and the finite size of elections met in practice are significant. 
As a second example, theoretical analysis is often
restricted to simple distributions. 
Manipulation may be very different in practice due to 
correlations between votes.
Walsh's empirical study 
was 
limited to the simple veto rule, weighted votes
and elections with only three candidates. In this paper, we relax
these assumptions and 
consider the more complex multi-round STV rule, 
unweighted votes,
and large numbers of 
candidates. 

Our experiments suggest that we should treat worst-case results
about the complexity of manipulating
voting rules like STV with some care. It was easy
for a single agent to manipulate almost every election in our 
experiments or to prove
that manipulation by a single agent was impossible. 
In the millions of elections
studied, we computed a manipulation or
proved none exists in a few minutes in each case. 
Indeed, in most cases, we took just a few seconds. 
As a result, 
we conjecture that it may be easy for a single agent to
compute a manipulation or prove none exists for
any STV election involving
a hundred or fewer agents and candidates. 
Since it may be unreasonable or infeasible
for an agent to order totally more candidates, and 
since elections with more agents are usually 
not manipulable by a single agent, 
this suggests that manipulation
of the STV rule by a single agent
is unlikely to be computationally 
hard in practice. 


\section{Manipulating STV}

Single Transferable Vote
(STV) proceeds in a number of rounds.
We consider the case of electing a single winner.
Each agent totally ranks the candidates. 
Unless one candidate has a majority of
first place votes, we eliminate the candidate
with the least number of first place votes. 
Any ballots placing the eliminated candidate in first 
place are re-assigned to the second place candidate.
We then repeat until one candidate has a majority. 
STV is used in a wide variety of elections
including for the Irish presidency, the
Australian House of Representatives, the Academy awards,
and many societies and organizations including the 
American Political Science Association,
the International Olympic Committee, 
and the British Labour Party. 

As in previous studies, we assume that the
agents manipulating the election know all the other votes.
The manipulation problem is to decide how 
the manipulators should vote, perhaps differently
to their true preferences, in order for a chosen
candidate to win. 
Amongst voting rules where the winner is polynomial
to compute, STV appears to be one of the
harder rules to compute how to manipulate. 
For example, 
it is NP-hard to manipulate
by a single voter if the number of candidates is
unbounded and votes are unweighted
\cite{stvhard}, or by a coalition of voters
if there are 3 or more candidates and
votes are weighted (or votes are
unweighted but there is uncertainty about 
the votes) \cite{csljacm07}.
Coleman and Teague
give an enumerative method
for a coalition of $k$ unweighted
voters to compute a
manipulation of the
STV rule which runs
in $O(m!(n+mk))$ time 
where $n$ is the number of voters and $m$ is
the number of candidates \cite{ctcats2007}.
For a single manipulator, 
Conitzer gave a method
to compute the set of candidates that a single
agent can make win a STV election
which takes $O(n 1.62^m)$ time
\cite{conitzerthesis}.
We improve upon Conitzer's procedure to compute
if a single agent can manipulate a
(possibly weighted)
STV election to make a desired candidate win. 
Unlike Conitzer's procedure, we ignore any
election in which the chosen candidate is eliminated,
and terminate search as soon as we find
any manipulation in which the chosen
candidate wins. 
 
\section{Uniform votes}

We start with one of the simplest possible
scenarios: elections in which each vote is equally
likely. We have one agent trying to manipulate an
election of $m$ candidates where $n$ other
agents vote. Votes are drawn uniformly
at random from all $m!$ possible votes. 
This is the Impartial Culture (IC) model. 

\subsection{Varying number of candidates}

\begin{figure}[htb]
\vspace{-2.5in}
\begin{center}
\includegraphics[scale=0.4]{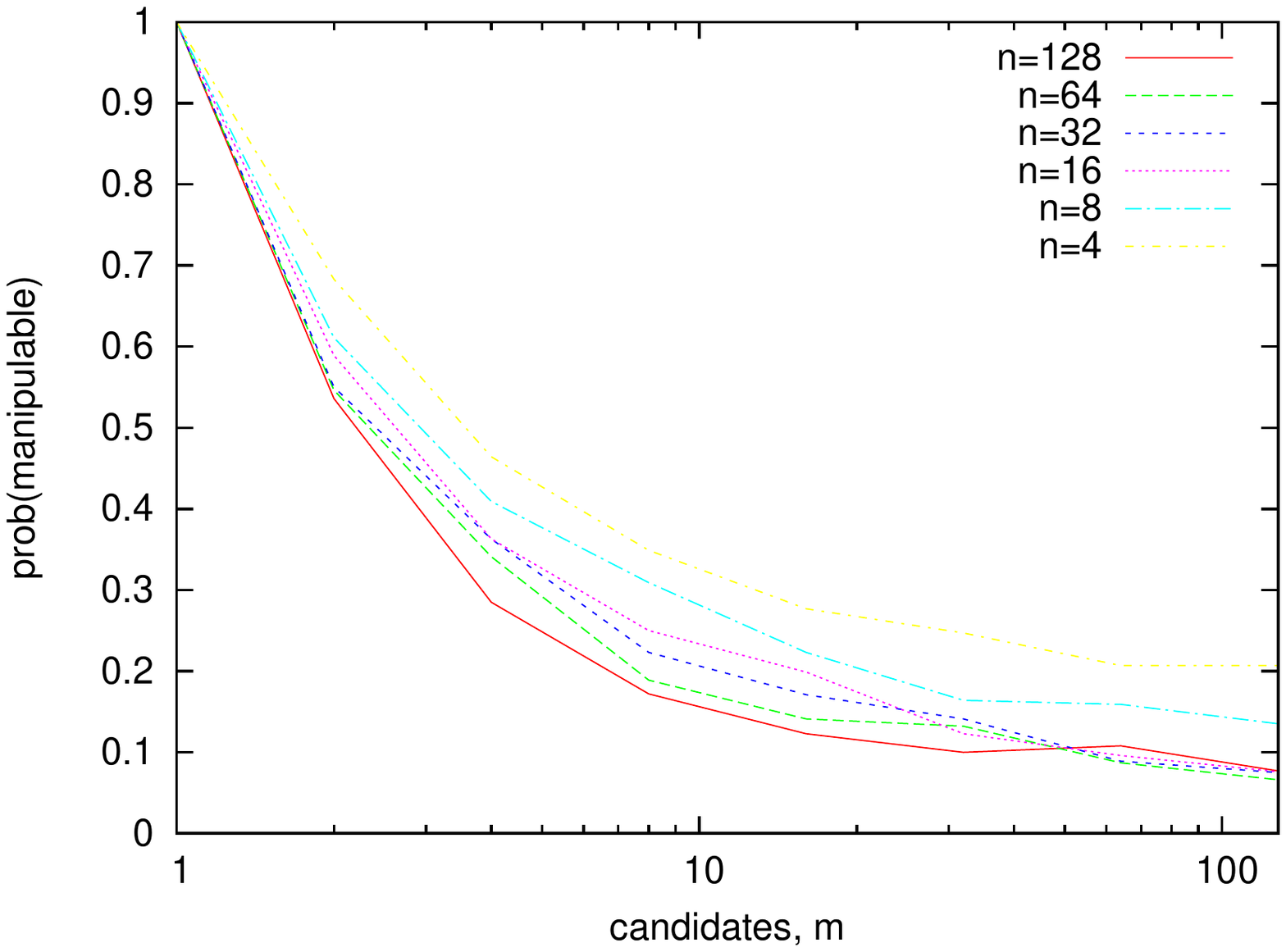}
\end{center}
\vspace{-0.5in}
\caption{Manipulability of 
random uniform voting. }
{\textnormal 
The number of agents voting is fixed and we vary the number of
candidates. The $n$ fixed votes are uniformly
drawn at random from the $m!$ possible votes. 
The y-axis measures the
probability that the manipulator can make a random
candidate win. }
\label{fig-prob-varm}
\end{figure}

\begin{figure}[htb]
\vspace{-2.5in}
\begin{center}
\includegraphics[scale=0.4]{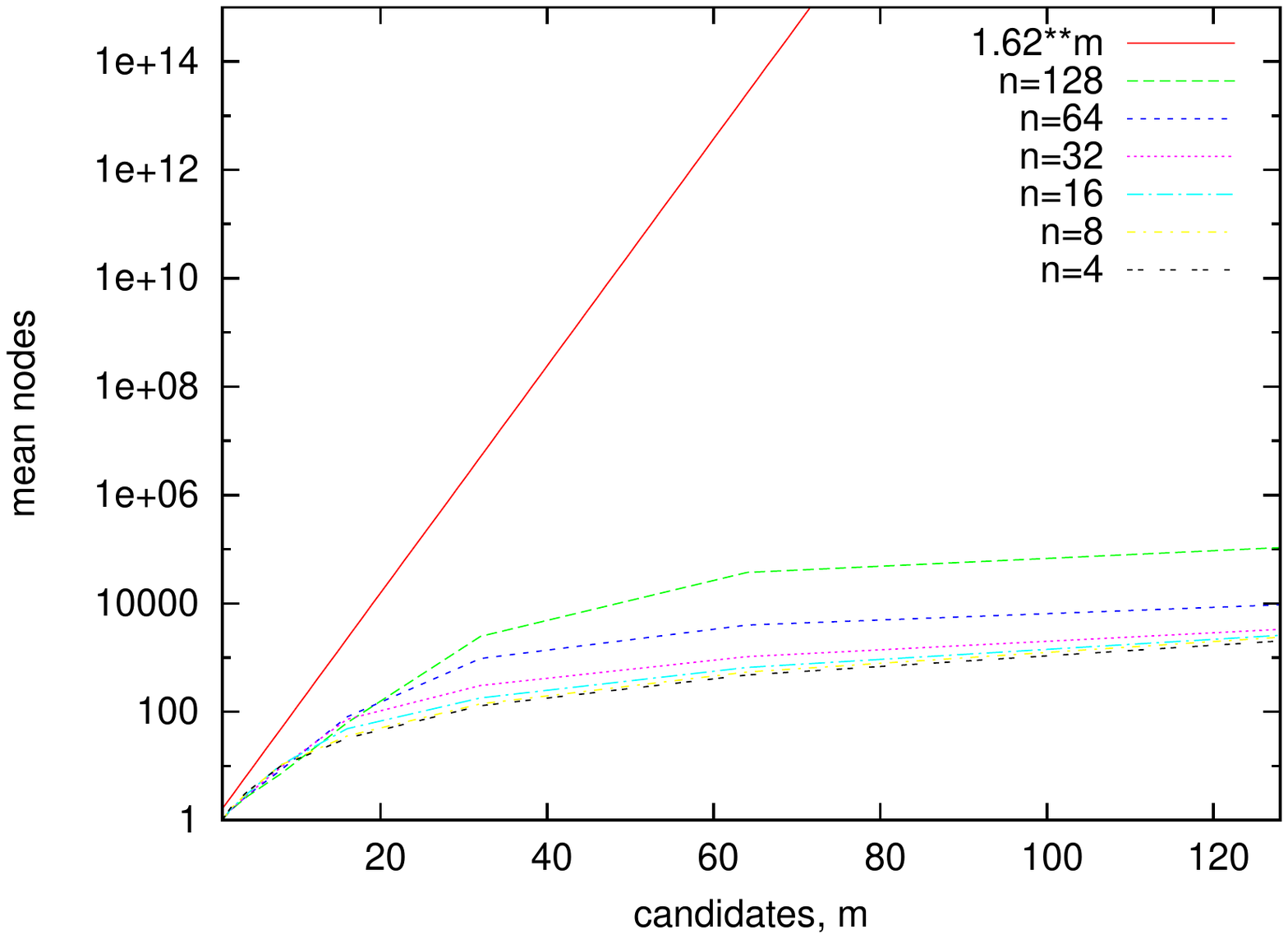}
\end{center}
\vspace{-0.5in}
\caption{Search to compute if an agent can manipulate an election
with random uniform voting.}
{The number of agents voting is fixed and we vary the number of
candidates. The y-axis measures the mean 
number of nodes explored to compute
a manipulation or prove that none exists. Median
and other percentiles are similar.}
\label{fig-nodes-varm}
\end{figure}

In Figures \ref{fig-prob-varm} and \ref{fig-nodes-varm},
we plot the probability that a manipulator can 
make a random agent win, and the search to compute
if this is possible when we 
fix the number of agents but vary the number of candidates. 
In this and subsequent experiments,
we tested 1000 problems at each point.
Unless otherwise indicated, 
the number of candidates $m$ and 
the number of agents $n$ are varied in powers of 2 from
1 to 128. 

As the number of candidates $m$ increases, the ability of an agent
to manipulate the election decreases. Unlike 
phase transition behaviour 
\cite{cheeseman-hard} in domains
like satisfiability \cite{mitchell-hard-easy,SAT-phase},
constraint satisfaction \cite{gmpwcp95,random},
number partitioning \cite{rnp,gw-ci98},
graph colouring \cite{wijcai99,wijcai2001} and
the traveling salesperson problem \cite{GentIP:tsppt}, 
the probability curve does not appear to sharpen to 
a step function around a fixed point. The probability
curve resembles the smooth phase transitions
seen in polynomial problems
like 2-coloring \cite{achlioptasphd} and 1-in-2 satisfiability
\cite{waaai2002}. 

Whilst the cost of computing a manipulation
increases exponential with the number
of candidates $m$, the observed scaling is much better than
the $1.62^m$. 
We can easily compute manipulations
for up to 128 candidates. There are perhaps few real world
elections with more candidates than this. 
Note that $1.62^m$ is over $10^{26}$ for $m=128$. Thus, we
appear to be far from the worst case. 
We fitted the observed data to the model $ab^m$ and found
a good fit with $b=1.008$ and a coefficient of determination, $R^2=0.95$. 


\subsection{Varying number of agents}

\begin{figure}[hptb]
\vspace{-2.5in}
\begin{center}
\includegraphics[scale=0.4]{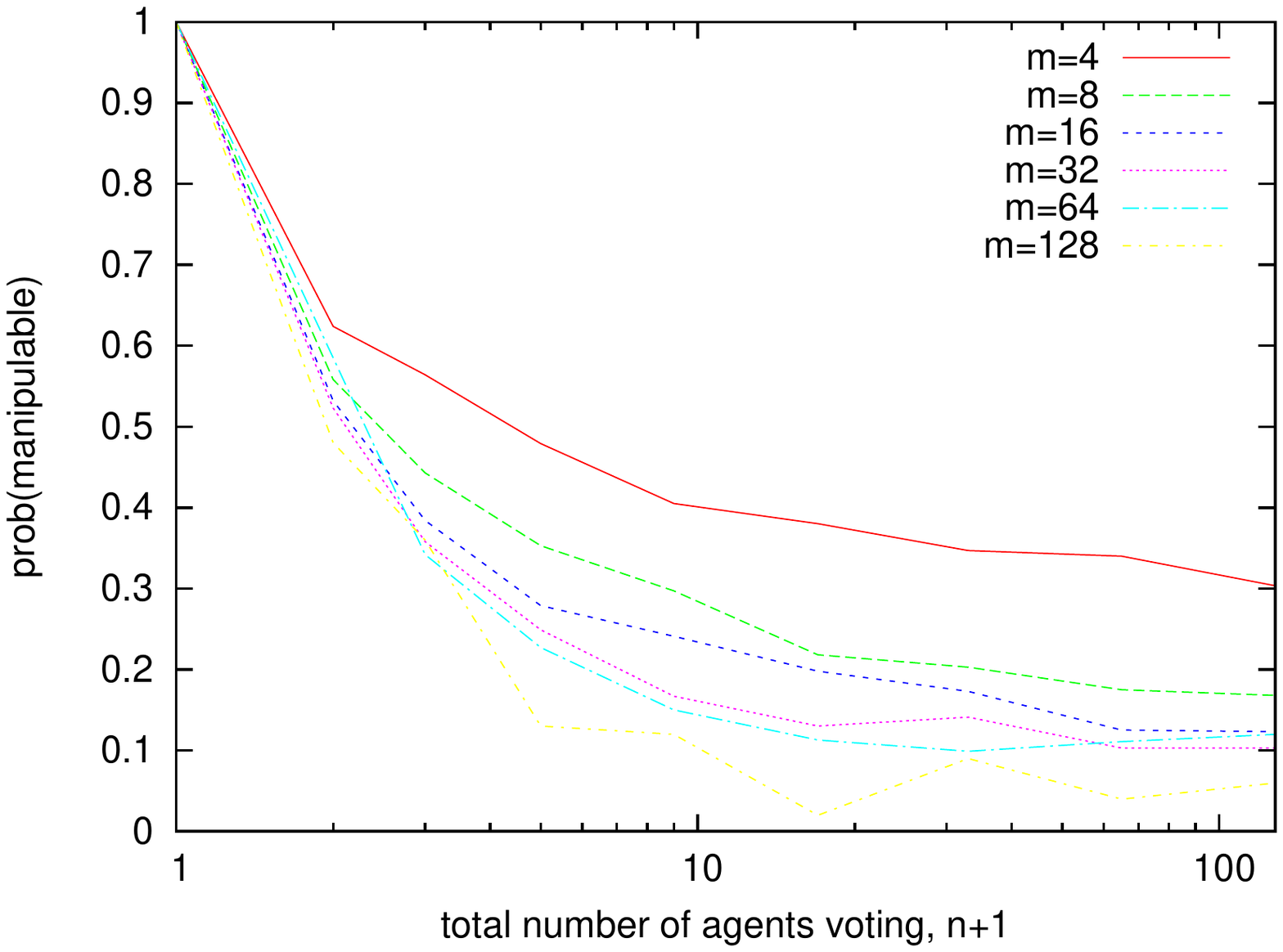}
\end{center}
\vspace{-0.5in}
\caption{Manipulability of random uniform votes.}
{The number of candidates is fixed and we vary the number of
agents voting. 
}
\label{fig-prob-varn}
\end{figure}

\begin{figure}[hptb]
\vspace{-2.5in}
\begin{center}
\includegraphics[scale=0.4]{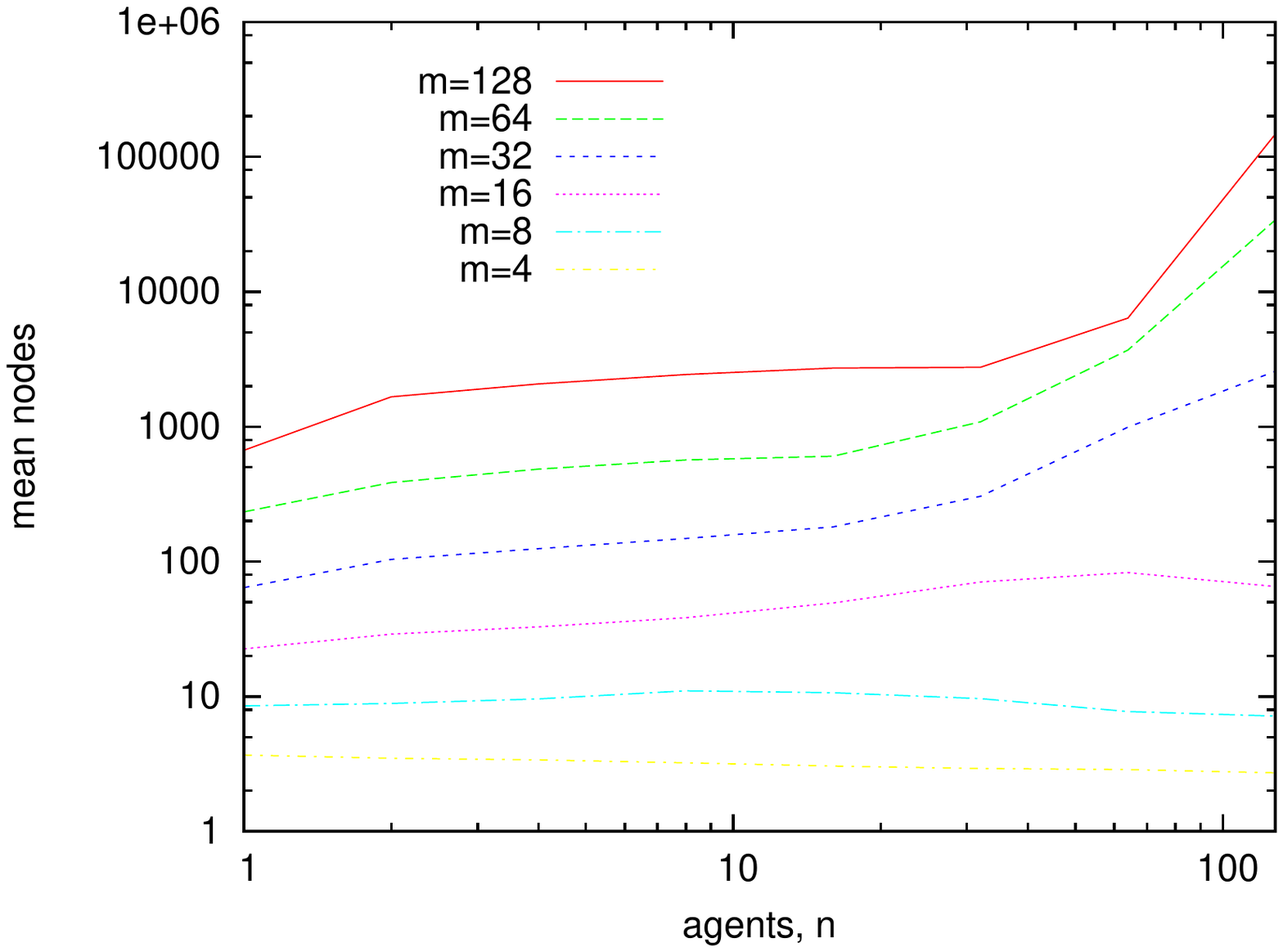}
\end{center}
\vspace{-0.5in}
\caption{Search to compute if an agent can manipulate an election
with random uniform votes.}
{The number of candidates is fixed and we vary the number of
agents voting. Median
and other percentiles are similar.}
\label{fig-nodes-varn}
\end{figure}

In Figures \ref{fig-prob-varn} and \ref{fig-nodes-varn},
we plot the probability that a manipulator can 
make a random agent win, and the cost to compute
if this is possible when we fix the number of
candidates but vary the number of agents
in the election. 
The ability of an agent
to manipulate the election decreases as the
number of agents, $n$ increases. 
Only if there are few votes and 
few candidates is there a significant chance
that the manipulator will be able to change
the result. 
Finding a manipulation or proving none
is possible is easy unless we have both a 
a large number of agents and a large number
of candidates. However, in this situation, 
the chance that the manipulator can
change the result is very small.

\section{Urn model}

To study the impact of more correlation between
votes, we considered random votes
drawn from the Polya Eggenberger urn model \cite{polya-urn}.
In this model, 
we have an urn containing all $m!$ possible votes. 
We draw votes out of the urn at random, and
put them back into the urn with $a$ 
additional votes of the same type (where
$a$ is a parameter).
As $a$ increases, there is 
increasing correlation between the votes. 
This generalizes both the Impartial Culture
model ($a=0$) and the Impartial Anonymous Culture ($a=1$) model. 
To give a parameter independent of problem 
size, we consider $b=\frac{a}{m!}$. 
For instance, with $b=1$, there is a 50\% chance
that the second vote is the same as the first. 

\begin{figure}[hptb]
\vspace{-2.5in}
\begin{center}
\includegraphics[scale=0.4]{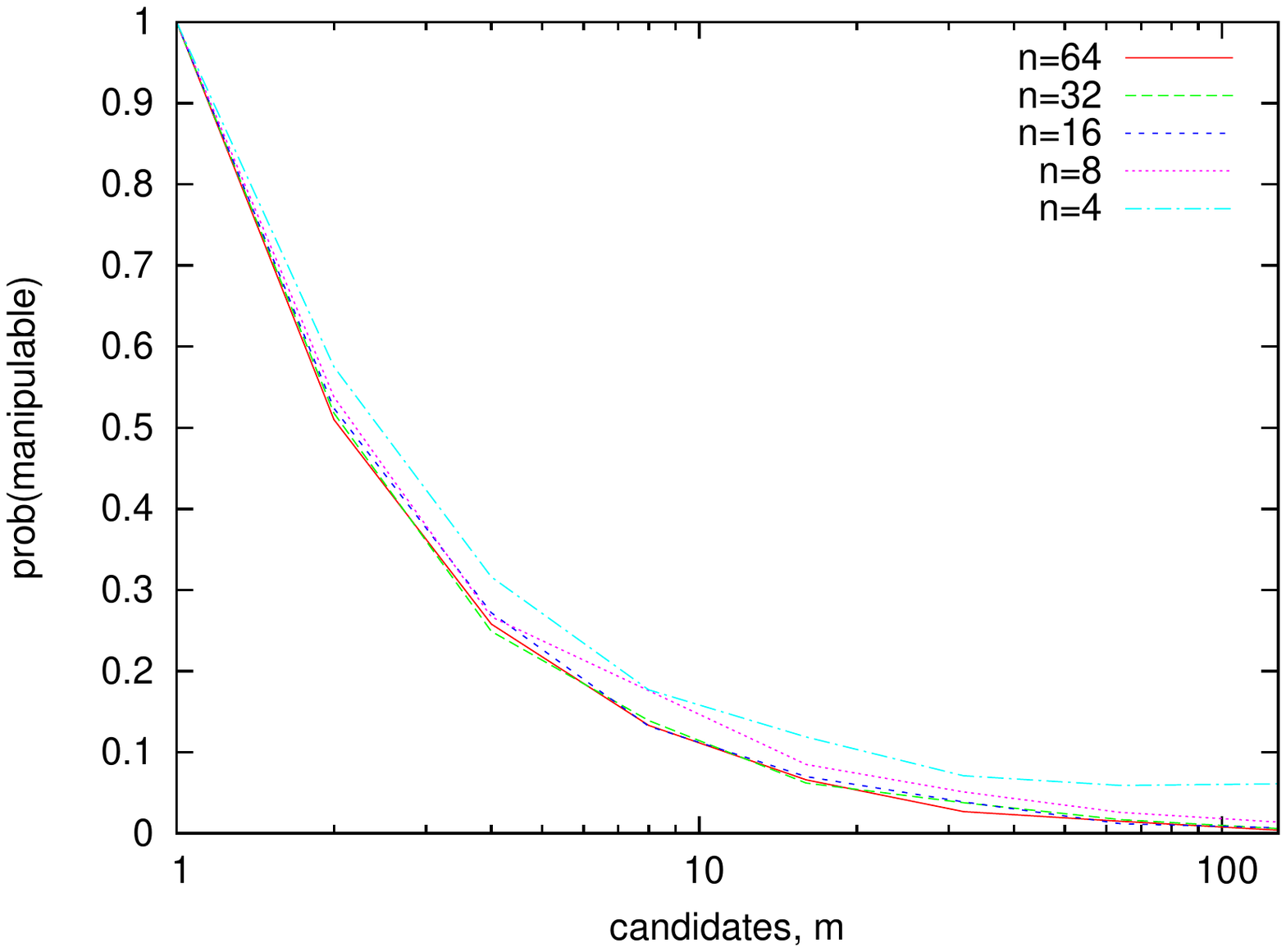}
\end{center}
\vspace{-0.5in}
\caption{Manipulability of 
correlated votes. }
{\textnormal 
The number of agents voting is fixed and we vary the number of
candidates. 
The $n$ fixed votes are drawn using using the Polya Eggenberger urn  model
with $b=1$. 
The y-axis measures the
probability that the manipulator can make a random
candidate win.
}

\label{fig-urn-prob-varm}
\end{figure}
\begin{figure}[hptb]
\vspace{-2.5in}
\begin{center}
\includegraphics[scale=0.4]{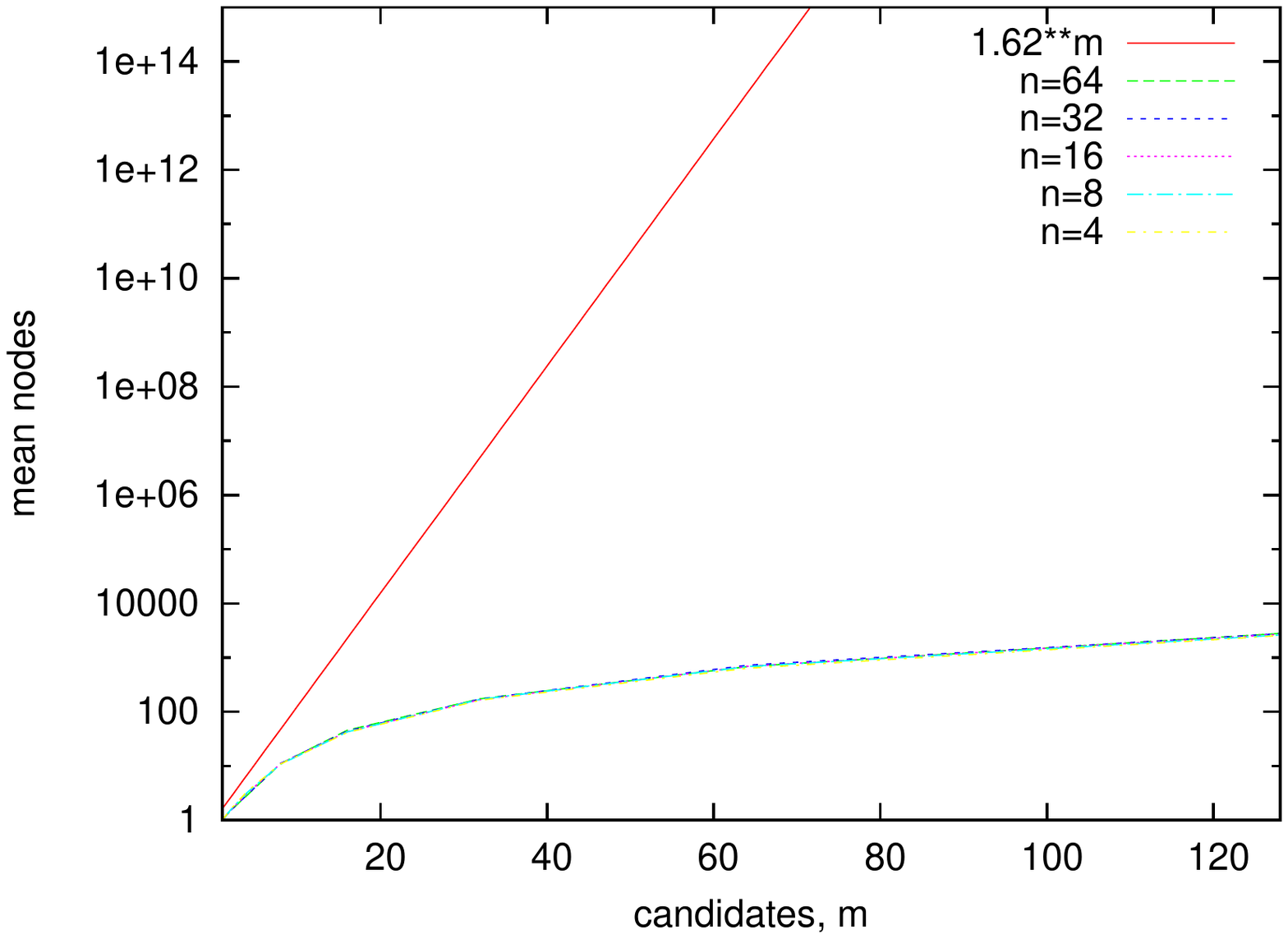}
\end{center}
\vspace{-0.5in}
\caption{Search to compute if an agent can manipulate an election
with correlated votes.}
{The number of agents voting is fixed and we vary the number of
candidates. 
The $n$ fixed votes are drawn using the Polya Eggenberger urn  model
with $b=1$. 
The y-axis measures the mean 
number of search nodes explored to compute
a manipulation or prove that none exists. 
The curves for different $n$ fit closely on top of each other.
Median
and other percentiles are similar.}
\label{fig-urn-nodes-varm}
\end{figure}

\begin{figure}[hptb]
\vspace{-2.5in}
\begin{center}
\includegraphics[scale=0.4]{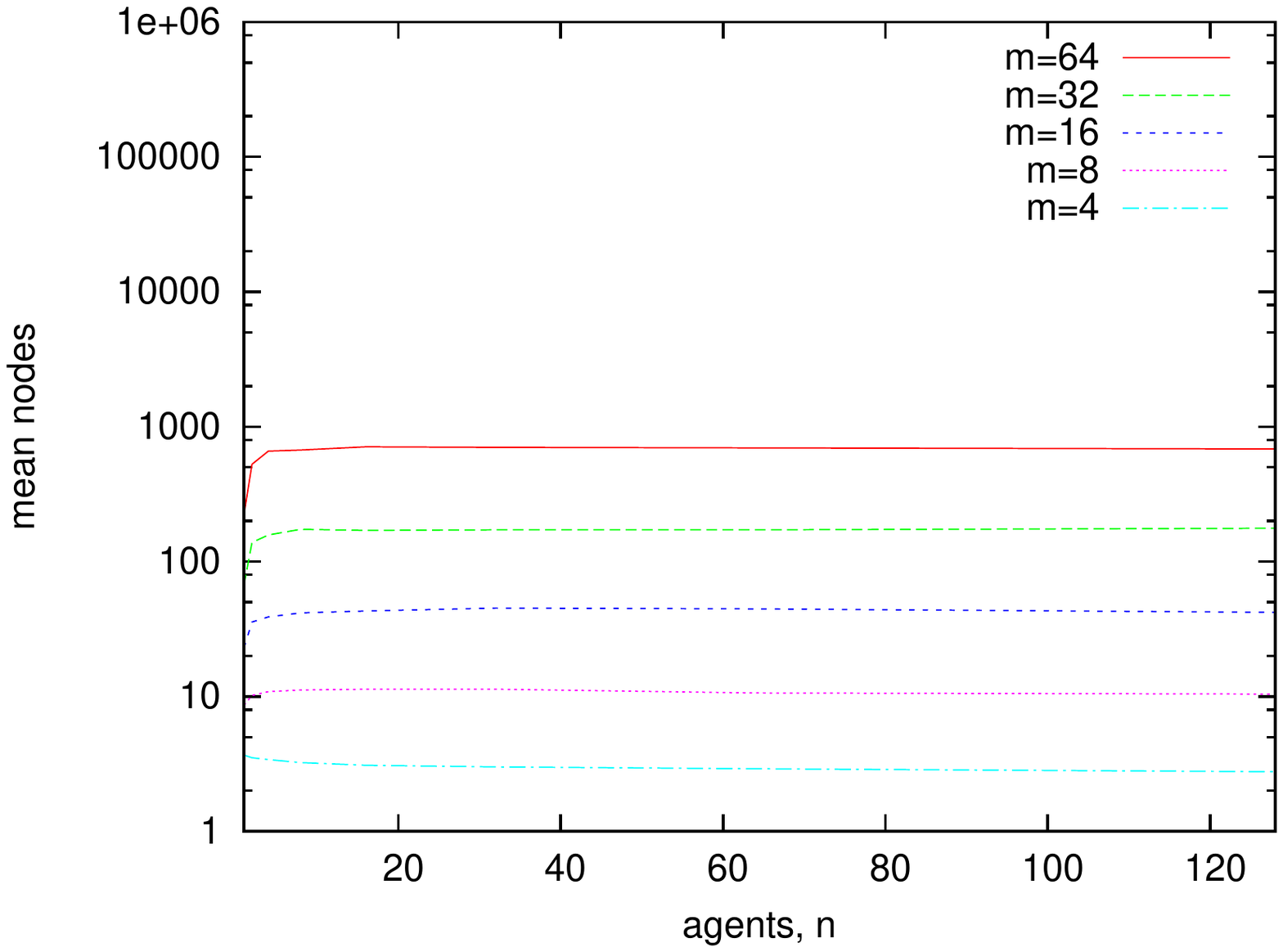}
\end{center}
\vspace{-0.5in}
\caption{Search to compute if an agent can manipulate an election
with correlated votes.}
{The number of candidates is fixed and we vary the number of
agents voting. 
The $n$ fixed votes are drawn using the Polya Eggenberger urn  model
with $b=1$. 
The y-axis measures the mean 
number of search nodes explored to compute 
a manipulation or prove that none exists. Median
and other percentiles are similar.}
\label{fig-urn-nodes-varn}
\end{figure}

In Figures \ref{fig-urn-prob-varm} and \ref{fig-urn-nodes-varm},
we plot the probability that a manipulator can 
make a random agent win, and the cost to compute
if this is possible as we vary the number of candidates
in an election where votes are 
drawn from the Polya Eggenberger urn model.
The search cost to compute a manipulation
increases exponential with the number
of candidates $m$. However, we can
easily compute manipulations
for up to 128 candidates and agents. 
We fitted the observed data to the model $ab^m$ and found
a good fit with $b=1.001$ and 
a coefficient of determination, $R^2=0.99$.

In Figure \ref{fig-urn-nodes-varn},
we plot the cost to compute a manipulation
when we fix the number of candidates
but vary the number of agents. 
As in previous experiments, finding a manipulation or proving none 
exists is easy even if we have many 
agents and candidates.

\section{Sampling real elections}

Elections met in practice may differ
from those sampled so far. There might, for instance, 
be some votes which are never cast. 
On the other hand, with the models studied so far every possible
vote has a non-zero probability of being seen. 
We therefore sampled some real voting
records \cite{isai95,ghpwaaai99}. 
\begin{figure}[hptb]
\vspace{-2.5in}
\begin{center}
\includegraphics[scale=0.4]{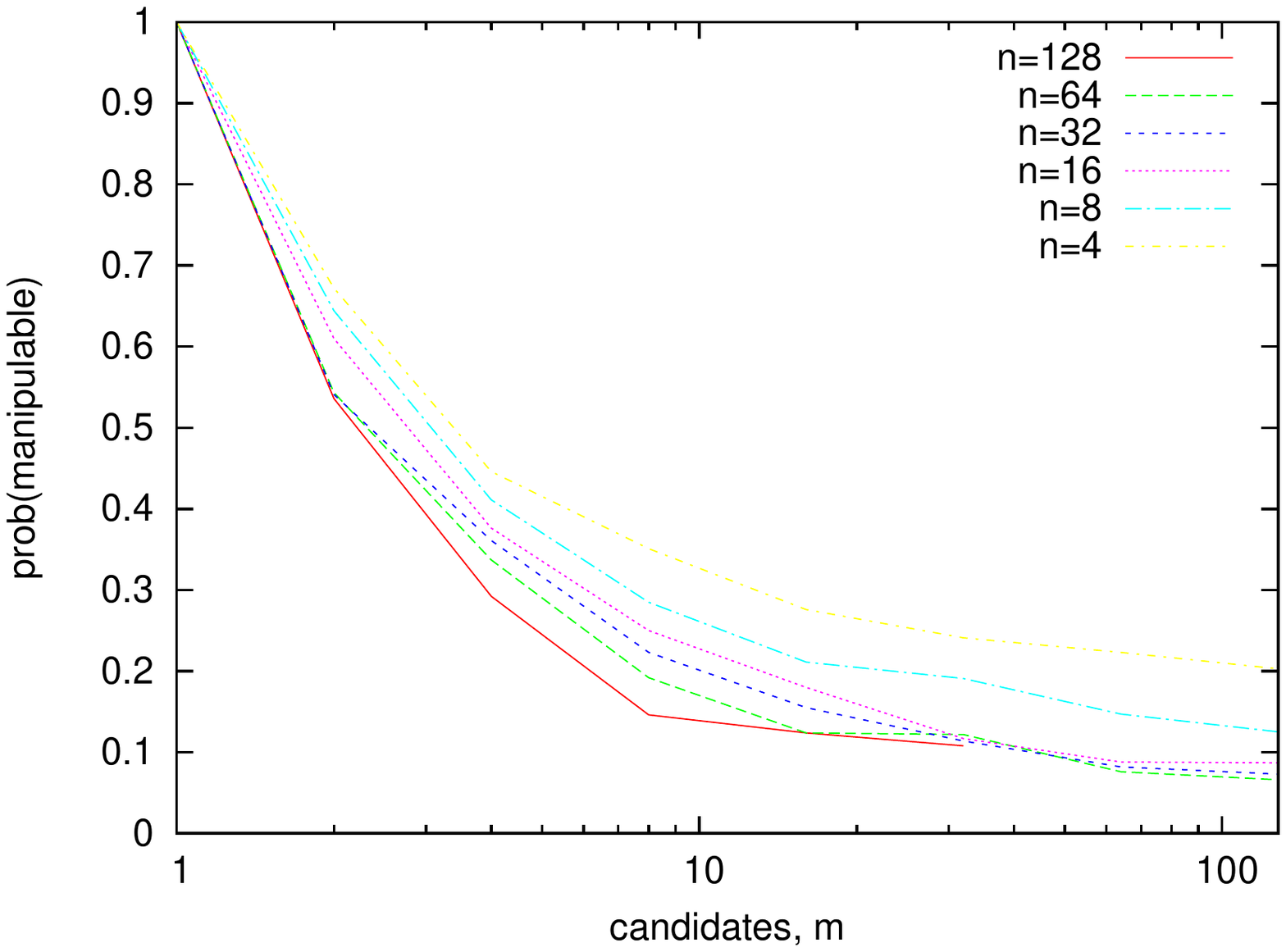}
\end{center}
\vspace{-0.5in}
\caption{Manipulability of 
votes sampled from the NASA experiment. }
{\textnormal 
The number of agents voting is fixed and we vary the number of
candidates. 
The y-axis measures the
probability that the manipulator can make a random
candidate win.
}
\label{fig-nasa-prob-varm}
\end{figure}
\begin{figure}[hptb]
\vspace{-2.5in}
\begin{center}
\includegraphics[scale=0.4]{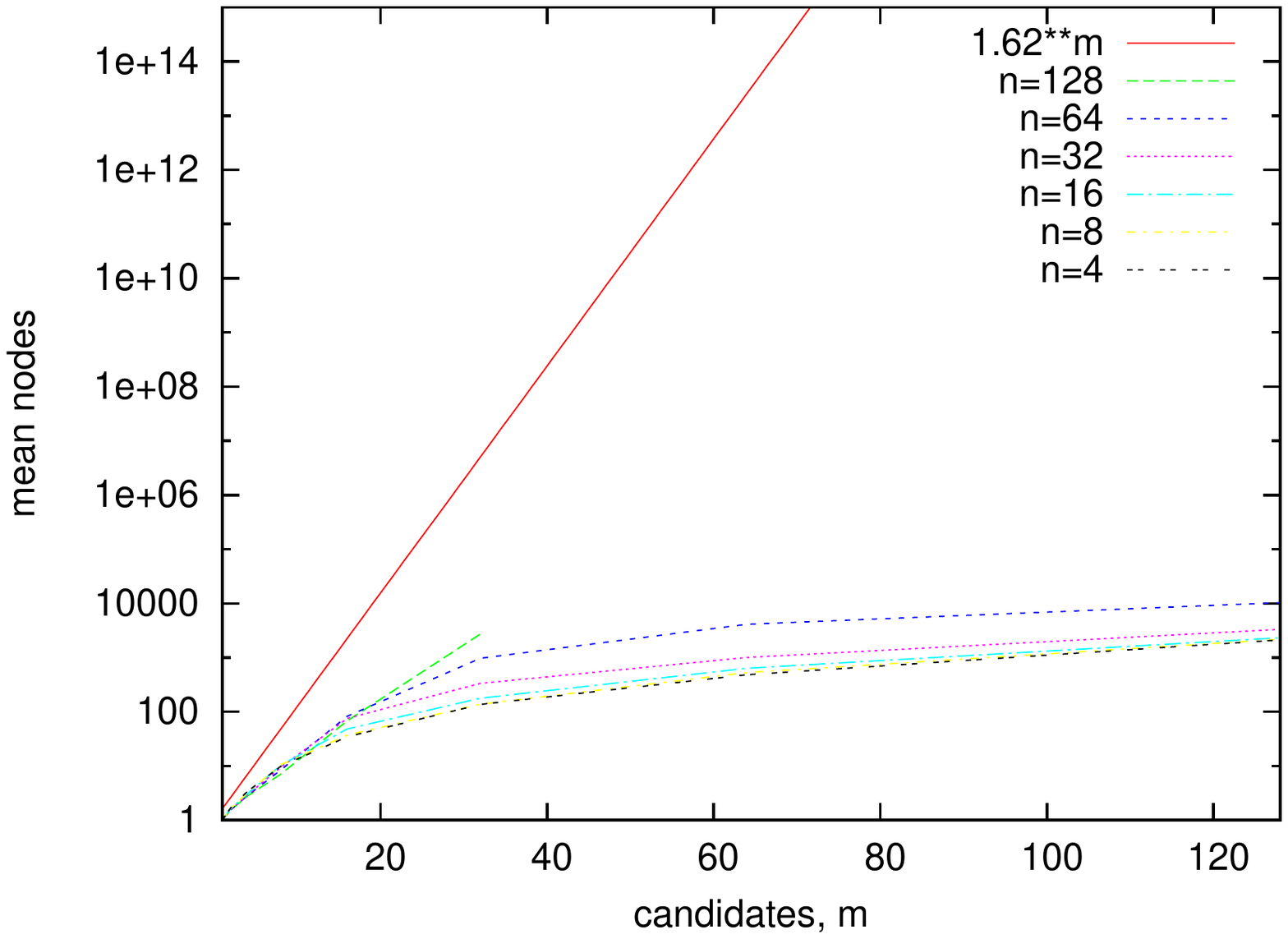}
\end{center}
\vspace{-0.5in}
\caption{Search to compute if an agent can manipulate an election
with votes sampled from the NASA experiment.}
{The number of agents voting is fixed and we vary the number of
candidates. 
The y-axis measures the mean 
number of search nodes explored to compute
a manipulation or prove that none exists. Median
and other percentiles are similar.}
\label{fig-nasa-nodes-varm}
\end{figure}
\begin{figure}[hptb]
\vspace{-2.5in}
\begin{center}
\includegraphics[scale=0.4]{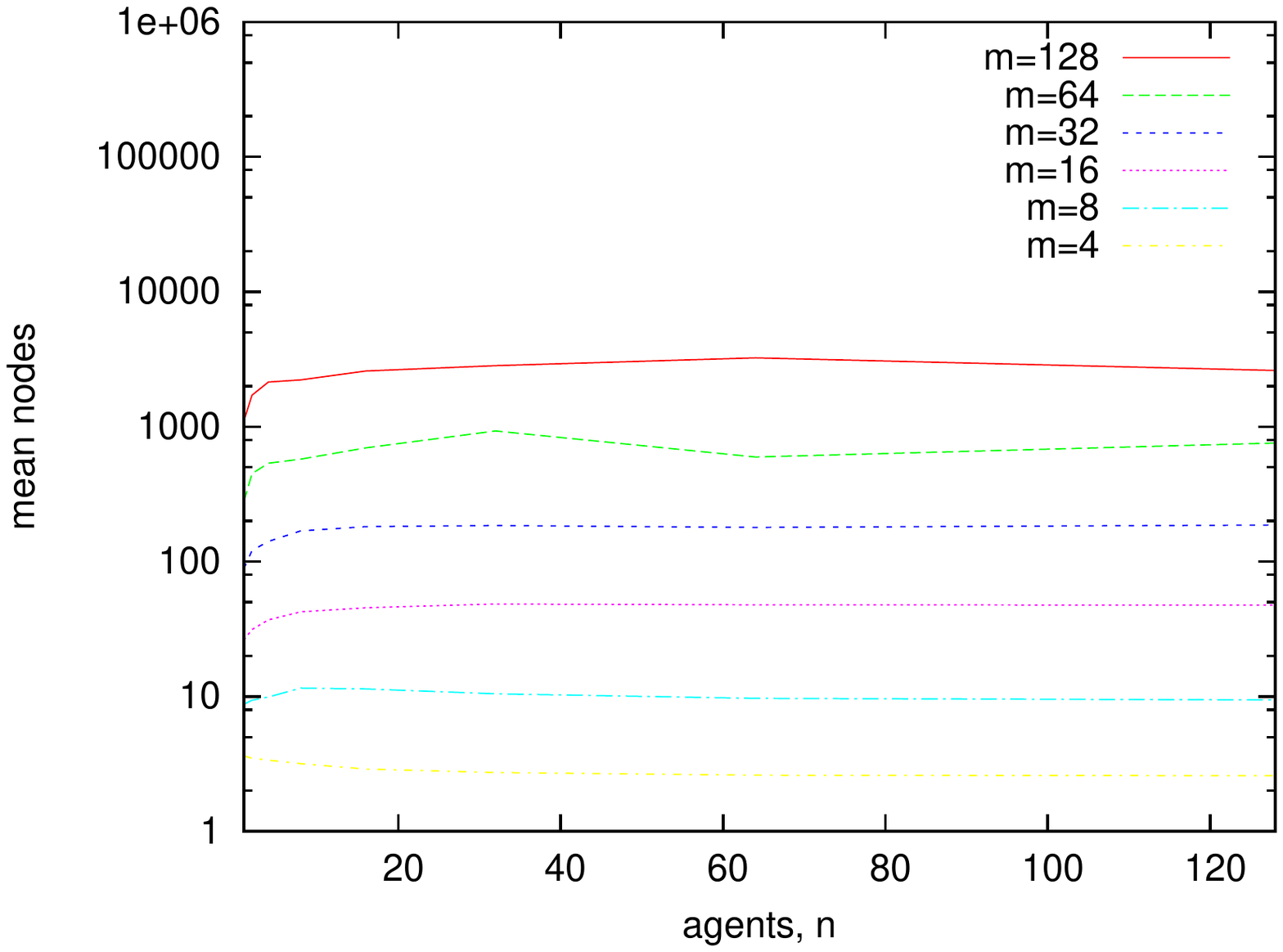}
\end{center}
\vspace{-0.5in}
\caption{Search to compute if an agent can manipulate an election
with votes sampled from the NASA experiment.}
{The number of candidates is fixed and we vary the number of
agents voting. 
The y-axis measures the mean 
number of search nodes explored to compute 
a manipulation or prove that none exists. Median
and other percentiles are similar.}
\label{fig-nasa-nodes-varn}
\end{figure}
Our first experiment uses the votes cast by 10 teams 
of scientists to select one of 32 different
trajectories for NASA's Mariner
spacecraft \cite{mariner}. Each team ranked the different
trajectories based on their scientific value. 
We sampled these votes. 
For elections with 10 or fewer agents voting,
we simply took a random subset of the 10 votes. 
For elections with more than 10 agents voting,
we simply sampled the 10 votes with uniform
frequency. For elections with 32 or
fewer candidates, we simply took a random subset of
the 32 candidates. Finally for elections with more than 32
candidates, we duplicated each candidate and assigned
them the same ranking. Since STV works on total
orders, we then forced each agent to break any ties
randomly.

In Figures \ref{fig-nasa-prob-varm} to \ref{fig-nasa-nodes-varn},
we plot the probability that a manipulator can 
make a random agent win, and the cost to compute
if this is possible as we vary the number of candidates
and agents.
Votes are sampled from the NASA experiment as explained
earlier. 
The search needed to compute a manipulation
again increases exponential with the number
of candidates $m$. However, 
the observed scaling is much better than
$1.62^m$. We can
easily compute manipulations
for up to 128 candidates and agents.

\begin{figure}[hptb]
\vspace{-2.5in}
\begin{center}
\includegraphics[scale=0.4]{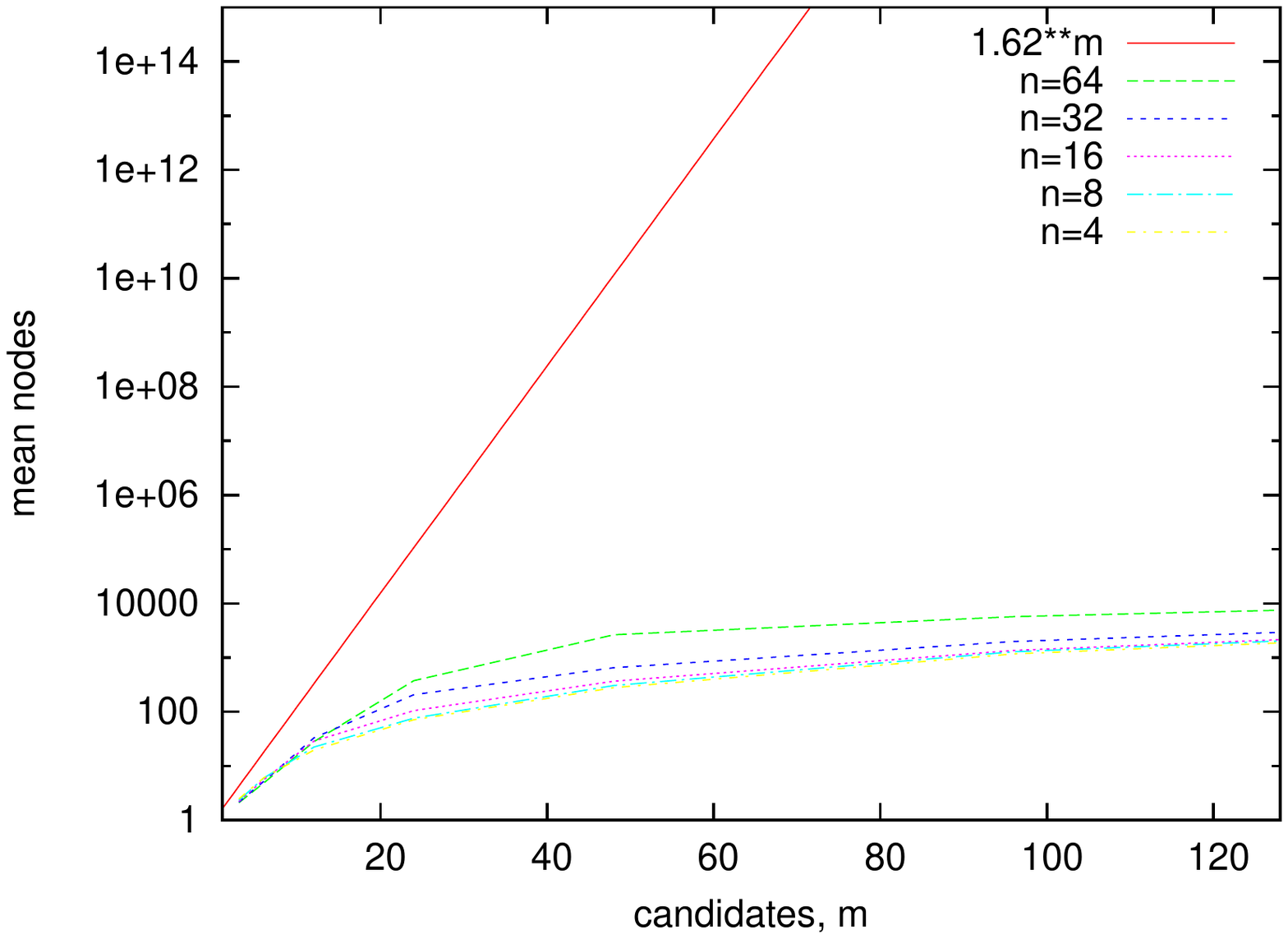}
\end{center}
\vspace{-0.5in}
\caption{Search to compute if an agent can manipulate an election
with votes sampled from a faculty hiring committee.}
{The number of agents voting is fixed and we vary the number of
candidates. 
The y-axis measures the mean 
number of search nodes explored to compute
a manipulation or prove that none exists. Median
and other percentiles are similar.}
\label{fig-dept-nodes-varm}
\end{figure}

\begin{figure}[hptb]
\vspace{-2.5in}
\begin{center}
\includegraphics[scale=0.4]{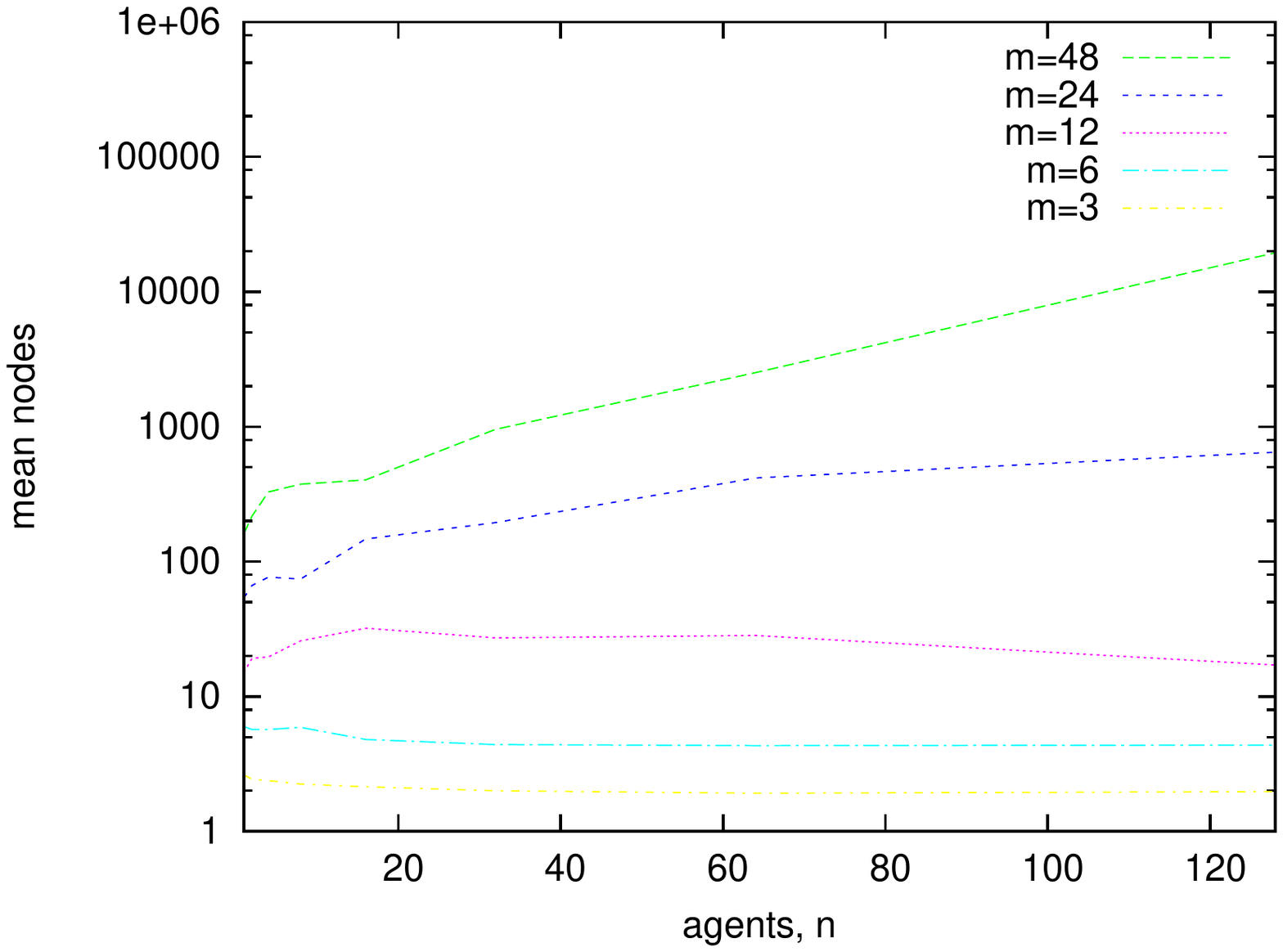}
\end{center}
\vspace{-0.5in}
\caption{Search to compute if an agent can manipulate an election
with votes sampled from a faculty hiring committee.}
{The number of candidates is fixed and we vary the number of
agents voting. 
The y-axis measures the mean 
number of search nodes explored to compute 
a manipulation or prove that none exists. Median
and other percentiles are similar.}
\label{fig-dept-nodes-varn}
\end{figure}

In our second experiment, we used
votes from a faculty hiring committee at the University
of California at Irvine \cite{dpc83}. 
This dataset had 10 votes for 3 different
candidates. We sampled from this 
in the same ways as from the NASA dataset.
Results are reported in Figures
\ref{fig-dept-nodes-varm} and
\ref{fig-dept-nodes-varn}. 
As in the previous experiments, 
it was easy to find a manipulation 
or prove that none exists. 
The observed scaling was again much better than
$1.62^m$.



\section{Related work}

Coleman and Teague proposed 
algorithms to compute a manipulation for
the STV rule \cite{ctcats2007}.
They also conducted an empirical study
which demonstrates that only relatively small
coalitions are needed to change the elimination
order of the STV rule. They observed that most uniform
and random elections
are not trivially manipulable using a simple greedy
heuristic. On the other hand, our results suggest 
that, for manipulation by a single agent, a limited amount of 
backtracking is needed to find
a manipulation or prove that none exists. 

Walsh empirically studied the cost
of manipulating the veto rule by
a coalition of agents whose votes 
were weighted \cite{wijcai09}. 
He showed that there was a smooth transition in the probability
that a coalition can 
elect a desired candidate as the size of 
the manipulating coalition increases.
He also showed that it was easy 
to find manipulations of the veto rule 
or prove that none exist 
for many independent and identically distributed votes
even when the
coalition of manipulators was critical in size. 
He was able to identify a situation
in which manipulation was computationally hard. 
This is when votes are highly correlated
and the election is ``hung''. 
However, even a single uncorrelated voter was enough to 
make manipulation easy again.

As indicated, there have been several theoretical
results recently that suggest elections are 
easy to manipulate in practice despite
NP-hardness results. 
For instance, Xia and Conitzer have shown that for a large class
of voting rules including STV, as the number of agents grows, either 
the probability that a coalition
can manipulate the result is very small (as the
coalition is too small), or the
probability that they can easily manipulate the result
to make any alternative
win is very large \cite{xcec08}. 
They left open only a small interval in the size
for the coalition for which the coalition is 
large enough to be able to manipulate but
not obviously large enough to be able to manipulate the
result easily. 

As a second example, 
Procaccia and Rosenschein proved that for most
scoring rules and a wide variety of
distributions over votes, 
when the size of the coalition is $o(\sqrt{n})$, the probability
that they can change the result 
tends to 0, and when it is $\omega(\sqrt{n})$, the probability
that they can manipulate the result
tends to 1 \cite{praamas2007}.
They also gave a simple
greedy procedure that will find a manipulation 
of a scoring rule in polynomial
time with a probability of failure that is 
an inverse polynomial in $n$
\cite{prjair07}.
Friedgut, Kalai and Nisan proved that
if the voting rule is neutral and
far from dictatorial and there
are 3 candidates then there exists
an agent for whom a random manipulation 
succeeds with probability $\Omega(\frac{1}{n})$
\cite{fknfocs09}. 
Starting from different assumptions, Xia and Conitzer showed that
a random
manipulation would succeed with probability
$\Omega(\frac{1}{n})$ for 3 or more 
candidates for STV, for 4 or more candidates for any scoring
rule and for 5 or more candidates for Copeland \cite{xcec08b}.

\section{Conclusions}

We have studied empirically
whether computational
complexity is a barrier to the manipulation
for the STV rule.
We have looked at a number of different
distributions of votes including
uniform random votes,
correlated votes drawn from an
urn model, and votes sampled from some
real world elections.
We have looked at manipulation 
by both a single agent, and a coalition
of agents who vote in unison.
Unlike phase transition behaviour
in domains like satisfiability,
we did not observe rapid transitions
in the manipulability of STV elections
that sharpen around a fixed point, but saw
instead smooth transitions. 
We also did not observe hard instances
around the transition in manipulability.
Indeed, almost every one of the millions of
elections in our 
experiments was easy to manipulate or to prove
could not be manipulated. 
Our results suggest that it may be easy for a single agent to
compute a manipulation or prove none is possible for
any STV election involving
a hundred or fewer of agents and candidates. 
It would also be interesting
to perform a similar empirical study
for other voting rules as well as for other types of
manipulation and control (e.g. manipulation
by a coalition, or control by the
addition/deletion of candidates). Two interesting
rules to study are maximin and ranked pairs. These
two rules have only recently been shown to be NP-hard
to manipulate, and
are members of the small set of voting rules which 
are NP-hard to manipulate without weights
or uncertainty 
\cite{xzpcrijcai09}. 



%

\end{document}